\pdfoutput=1

\documentclass[11pt]{article}

\usepackage[]{emnlp2021}

\usepackage{times}

\usepackage{latexsym}
\usepackage{tabularx}
\usepackage{arydshln}

\usepackage[T1]{fontenc}

\usepackage[utf8]{inputenc}

\usepackage{microtype}
\usepackage[normalem]{ulem}
\usepackage{booktabs}
\usepackage{threeparttable}
\usepackage{color,soul}
\usepackage{graphicx} 
\usepackage{tikz}
\usepackage{pgfplots}
\definecolor{DarkBlue}{RGB}{28,81,140}
\definecolor{DarkRed}{RGB}{151,11,28}
\definecolor{DarkGreen}{RGB}{49,147,70}
\definecolor{LightGreen}{RGB}{129,234,108}
\makeatletter
  \newcommand\greenuline{\bgroup\markoverwith{\textcolor{green}{\rule[-0.7ex]{2pt}{0.4pt}}}\ULon}
  \UL@protected\def\redwave{\leavevmode \bgroup 
    \ifdim \ULdepth=\maxdimen \ULdepth 3.5\p@
    \else \advance\ULdepth2\p@ 
    \fi \markoverwith{\lower\ULdepth\hbox{\textcolor{red}{\sixly \char58}}}\ULon}
\makeatother

\usepackage{times}
\usepackage{multicol}
\usepackage{multirow}
\usepackage{graphicx}
\usepackage{booktabs}
\usepackage{microtype}
\usepackage{algorithm}
\usepackage{algorithmic}
\usepackage{amsbsy}
\usepackage{amsmath,soul}
\usepackage{soulpos}
\usepackage{amsmath}
\usepackage{multirow}
\usepackage{makecell}

\usepackage{amsmath,soul}
\usepackage{comment}
\usepackage{soulpos}
\usepackage{xcolor}
\usepackage[english]{babel}
\usepackage{blindtext}
\usepackage[shortcuts]{extdash}
\ulposdef{\ulnumaux}{%
   $\underset{\saveulnum}{\rule[-.7ex]{\ulwidth}{.4pt}}$}

\newcommand{\ulnum}[2]{%
  \def\saveulnum{#1}%
  \ulnumaux{#2}}
\DeclareRobustCommand{\cmss}[1]{{{\fontfamily{cmss}\selectfont{#1}}}}

\newcommand{\mich}[1]{{\textcolor{violet}{#1}}}
\newcommand{\ehsan}[1]{{\textcolor{cyan}{#1}}}

\newcommand\boldorange[1]{\textcolor{orange}{\textbf{#1}}}

\title{It is Not as Good as You Think!\\
 {E}valuating Simultaneous Machine Translation on Interpretation Data}

\author{Jinming Zhao$^1$\ \   Philip Arthur$^2$\ \  Gholamreza Haffari$^1$\ \  Trevor Cohn$^3$\ \  Ehsan Shareghi$^{1,4}$ \\
$^1$Department of Data Science \& AI, Monash University \\
$^2$Oracle Digital Assistant, Oracle Corp.\ \ \ $^4$Language Technology Lab, University of Cambridge\\ 
$^3$School of Computing and Information Systems, The University of Melbourne \\
{first.last}@\{monash.edu, unimelb.edu.au, oracle.com\}}

\begin{document}
\maketitle

\begin{abstract}
Most existing simultaneous machine translation (SiMT) systems are trained and evaluated on offline translation corpora. We argue that SiMT systems should be trained and tested on real interpretation data. To illustrate this argument, we propose an interpretation test set and conduct a realistic evaluation of SiMT trained on offline translations. Our results, on our test set along with 3 existing smaller scale language pairs, highlight the difference of up-to 13.83 BLEU score when SiMT models are evaluated on translation vs interpretation data. In the absence of interpretation training data, we propose a translation-to-interpretation (T2I) style transfer method which allows converting existing offline translations into interpretation-style data, leading to up-to 2.8 BLEU improvement. However, the evaluation gap remains notable, calling for constructing large-scale interpretation corpora better suited for evaluating and developing SiMT systems.
\footnote{Our annotated test sets are available at \url{https://github.com/mingzi151/InterpretationData}.}

\end{abstract}


\section{Introduction}
\label{sec:introduction}

Simultaneous interpretation (SI) is a task of translating natural language in real time. 
SiMT systems are expected to generate interpreted text as if the text was produced by human interpreters while maintaining acceptable delay \citep{DBLP:conf/acl/MaHXZLZZHLLWW19, DBLP:conf/eacl/ArthurCH21}. However, most current SiMT systems are trained and evaluated on offline translations differing from real-life SI scenarios where
translations are flexibly paraphrased,
without compromising the source message~\citep{DBLP:conf/naacl/HeBD16, DBLP:conf/asru/PaulikW09}. For instance, in Table~\ref{tab:intro_example} the interpretation sentence drops "at this point" and condenses "seriousness of this line of argument" to "agreement"; it delivers the source message as reliably as the offline translation. 
\begin{table}[t]
    \small
   \scalebox{0.85}{
    \begin{tabular}{|p{8cm}|}
    \hline
        \textit{\textbf{Source:}}
        \textcolor{violet}{\ulnum{I'm}{Ich werde} \ulnum{at\ this\ point}{an diesem Punkt} \ulnum{refrain\ from}{darauf verzichten}, \ulnum{}{einen} \ulnum{to\ comment\ on\ the\
seriousness\ of\ this\ line\ of\ reasoning}{Kommentar zur  Ernsthaftigkeit dieser Argumentationsweise} \ulnum{}{abzugeben}}.  \\
        \textit{\textbf{Offline Translation:}} \textcolor{DarkBlue}{\boldorange{(}At this point,\boldorange{)} I will refrain from commenting on the \redwave{seriousness of this line of argument}.}  \\ 
        \textit{\textbf{Interpretation:}} \textcolor{DarkGreen}{I'm not going to comment on that \redwave{agreement}.} \\
    \hline
    \end{tabular}}
    \caption{
    Translation and interpretation differ in 
    style while conveying the same source information.
    }
    \vspace{-2.5mm}
    \label{tab:intro_example}
\end{table}

Prior work attempted to build interpretation corpora in a small scale \citep{tohyama2004ciair,DBLP:conf/lrec/ShimizuNSTN14,bernardini2016epic}, or constructed speech interpretation training corpora for MT tasks \citep{DBLP:conf/icassp/PaulikW10}. 
But, very little attempt has been made on empirically quantifying the evaluation gap.
%
%
%
An exception is \citet{shimizu2013constructing} which incorporated interpretation data in the training stage of a statistical MT system, but the lack of training data and the scale of evaluation set resulted in a marginal BLEU score difference.
\footnote{Concurrently, \citet{zhang2021bstc} trained a system on an offline corpus and evaluated on interpretation test sets, not available to the public at the time of writing our paper.}


We compile a genuine interpretation test set of 1k {utterances} from the European Parliament (EP) Plenary focusing on German$\rightarrow$English. We examine the real performance gap of wait-k~\citep{DBLP:conf/acl/MaHXZLZZHLLWW19}, a state-of-the-art SiMT system, on our test set along with 3 smaller scale~\citep{bernardini2016epic} translation and interpretation language-pairs and observe a drop of  up-to 13.83 BLEU score. In the absence of interpretation-style training data, we propose a simple and effective translation-to-interpretation (T2I) style transfer method to produce pseudo-interpretations from abundant offline translations. 
Training on our T2I transferred data, we
observe an improvement of $\sim$2.8 BLEU score.
%
Our findings necessitate further developments towards  constructing large-scale interpretation corpora, designing domain adaptive techniques and models more reflective of real-life interpretations. 

\section{German$\rightarrow$English Interpretation Data}
We provide an overview of our
data construction and move full details in \emph{Appendix A.1}.

\paragraph{Collection.}
We crawled data from the EP Plenary\footnote{\url{www.europarl.europa.eu/plenary/en/debates-video.html}} between 2008 and 2012\footnote{Beyond this period, offline translations are not provided.} and  downloaded 238 debates consisting of speech transcriptions,
offline translations and interpretation videos. We used Google speech API to transcribe the interpretation videos and normalize automatic speech recognition (ASR) outputs, yielding 323-hour of transcriptions. 

\paragraph{Cleaning, Alignment, and Segmentation.}
We removed duplicates and the dialogues with non-German source sentences, while using available offline translations to retrieve named entities; this resulted in 5,239 dialogues. We filtered out dialogues with interpretations less than 4 words, and call the resulting interpretations \textsf{Raw} hereafter. 

We further removed cases whose sources contained either (1) less than 20 tokens, (2) less than 150 words and included pre-defined signals, or (3) a different number of sentences from the corresponding offline translations. 
and whose sources and offline translations had a different number of sentences. Next a manual process was applied, including removals of dialogues with non-essential contents and truncation of interpretations whose first and last sentences did not match the corresponding offline translations (mostly due to imperfect audio segmentation). 987 dialogues\footnote{One dialogue is attached in \emph{Appendix A.1}. Note each source sentence and offline translation in the dialogues may consist of several sentences.} were thus retained, each of which having 14.5 sentences on average.

We aligned translations with transcriptions (interpretations). For each dialogue, as the transcriptions may not be well segmented in the ASR process, we identified sentences in the transcriptions with stanza \cite{qi2020stanza}, before segmenting them using dynamic programming. Manual inspection revealed that there were a portion of mismatched pairs, which was due to occasional interpreting failure resulting from interpreters' accumulated cognitive load \cite{mizuno2017simultaneous,DBLP:journals/corr/abs-2011-04845}. We further removed pairs the lengths of whose source and target were far off, and call it \textsf{Clean}, containing triples <\cmss{source, translation, interpretation}>.



\paragraph{Translation and Interpretation Test Sets.} To ensure the quality of interpretation data for evaluation, we hired a bilingual German-English speaker to annotate a randomly selected subset (107 dialogues) of the 987 dialogues in two stages: segmentation and ASR error correction. This gave us two versions of test set: \textsf{Interpretation$^{ASR}$}, \textsf{Interpretation}. 

In the first stage, the annotator was asked to match the correct target sentence(s) against each source sentence. The annotator was asked to find interpretation text for each German sentence, when impossible, multiple sentences were allowed. Additionally, to comply with human speaking styles, we allowed minor omissions of unimportant English texts as long as the main idea of German text was conveyed (such as conjunctions).  In the second stage, the annotator was instructed to correct ASR errors while applying minimal changes to the sentences.

Ultimately, our test sets comprise 1,090 triples of \cmss{<source, translation, interpretation>} which were further cross checked to enforce quality control. 

\section{T2I Style Transfer}

 \begin{figure}[t]
  \centering
  \includegraphics[width=1\linewidth]{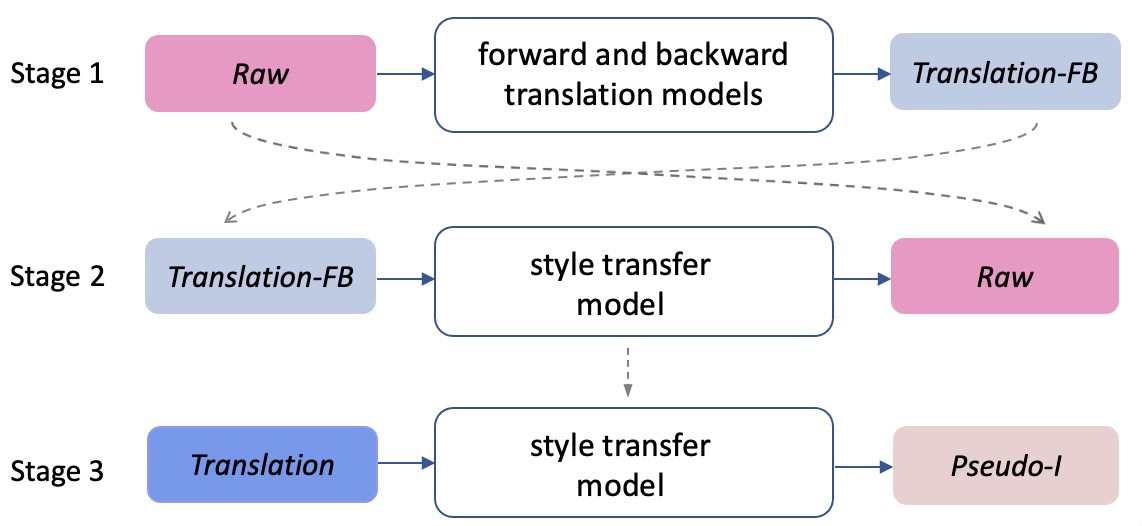}
  \caption{T2I style transfer in unsupervised settings.}
  \label{fig:res}
\end{figure}
Offline translated texts and online interpreted texts differ in various aspects, including lengths, sentence structure and lexicon; this is fundamentally contributed by the fact that interpreters use tactics to minimize delay and reduce the load of retention~\citep{mizuno2017simultaneous,camayd2011cognitive}. For example, interpreters tend to break a source sentence into several smaller chunks~(see \citet{ DBLP:conf/naacl/HeBD16} for more tactics).
%
%
Yet, while exhibiting stylistic differences, both preserve the key source message. As will be seen (\S\ref{sec:evalgap}) these differences amount to a significant evaluation gap.

While the ideal solution is using human annotators to create interpretation training corpora, in the absence of resources, we propose a simple technique, T2I style transfer, to convert existing translation data into interpretation-style data with a style transfer model. 
Training such a model would require paired translations and interpretations, which are not available in large quantities. Rather, our proposed approach allows fulfilling the goal of style transferring abundant translation data to interpretation-like data both in supervised settings, where \textsf{Clean} is leveraged, and unsupervised settings, where only \textsf{Raw} is used.

\paragraph{Supervised training} 
Given that our \textsf{Clean} set consists of roughly 4.2k triples, we opt for statistical MT systems which inherently require far less data for  sequence-to-sequence mapping tasks compared to their neural counterparts. Furthermore, conducting style transfer in the same language involves word replacement and ordering, which conforms with the behaviors of SMT systems  that chunk an input sequence into segments, translate, and reorder the translated chunks \cite{lopez2008statistical}.  More specifically, we employ two classic statistical MT methods: phrase-based SMT (PBMT) \citep{DBLP:conf/naacl/KoehnOM03} and Hierarchical phrase-based MT (HPBMT) \citep{DBLP:conf/acl/Chiang05}.\footnote{PBMT creates a phrase table, a reordering model and a language model, followed by tuning their weights with MERT on parallel data. HPBMT leverages both phrase-based translation and syntax-based translation, and operates on context-free grammar rules. }  A similar framework was tried by \citet{DBLP:conf/coling/XuRDGC12} for text simplification.

{We will describe the T2I pipeline process for unsupervised settings, as both settings have a similar process with different data configurations. The main difference is that we use \textsf{Clean} instead of \textsf{Raw}, which will be detailed in \S\ref{sec:data}.}

\paragraph{Unsupervised training} Figure \ref{fig:res} shows the three stages of our T2I approach in unsupervised settings: the first stage is to convert interpretations in \textsf{Raw} to translation-style data by applying round-trip translation on interpretations, with pretrained NMT models~\citep{DBLP:conf/wmt/NgYBOAE19}. 
It is expected that the outputs after this round-tripping, denoted as  \textsf{Translation-FB},  sit close to the translation domain, thus achieving the effects of interpretation-to-translation.
The second stage is to train a style transfer model to learn the mapping between the data points in \textsf{Translation\=/FB} and their corresponding interpretations in \textsf{Raw}. Lastly, we apply the trained style transfer model on offline Europarl translations and produce interpretation-like sequences which we call \textsf{Pseudo-I}.\footnote{Examples of \textsf{Pseudo-I} are provided in \emph{Appendix A.2}.}  

\section{Experiments}
\label{sec:experiments}

In this section, we present datasets details (\S\ref{sec:data}) followed by the descriptions of our baselines and style transfer models (\S\ref{sec:basemodels}). We report results by underlining the performance gap between evaluation on translated and interpreted texts (\S\ref{sec:evalgap}), and showing the effectiveness of our T2I style transfer both quantitatively and qualitatively (\S\ref{sec:t2ianalysis}).  

We followed the instructions in \citet{DBLP:conf/eacl/ArthurCH21} to preprocess data, and their hyperparameters for training all wait-k models. For style transfer models, we used the standard setup for both PBMT and HPBMT.\footnote{http://www.statmt.org/moses/?n=Moses.Overview}

\begin{table*}[!t]
\centering
\small
  \scalebox{1}
  {
  \begin{tabular}{lccccccccc}
    \toprule
    &\multicolumn{3}{c}{\# of pairs}&\multicolumn{6}{c}{Evaluation}\\ 
    \cmidrule(l){2-4}\cmidrule(l){5-10}
   &\multicolumn{2}{c}{Europarl Offline}&$\star$&\multicolumn{3}{c}{Translation Test}  &
    \multicolumn{3}{c}{Interpretation Test} \\
    \cmidrule(l){2-3}\cmidrule(l){5-7}\cmidrule(l){8-10}
    Lang.&Train&Dev&Test& AP & AL & Bleu  & AP & AL& Bleu\\
     \midrule
    DE &1,666,904 & 3,587&1,051$^+$& 0.61 & 2.84 & 22.78 & 0.61 & 2.84 & 12.34 \\ 
    FR &1,929,486 &9,736 & 675 & 0.58&2.41 &21.24 &0.58  &2.41 &9.28   \\
    PL &601,021&2,035&463 & 0.61& 2.94&24.24 &0.61  &2.94 &13.71 \\
    IT &1,832,809&9,256& 480& 0.56 &2.45 &24.47 & 0.56 &2.45 &10.64 \\
    \bottomrule
  \end{tabular}
  }

  \caption{Data statistics (\# pairs) and evaluation gap using translation vs interpretation test set. ($\star$) Test data for both  translation and interpretation sets for FR, PL, and IT were from EPTIC, and DE was our proposed set. ($+$) We further removed 39 cases from our 1090 triples which heavily overlapped with the training data. All training data were Europarl offline translations.}
  \label{tab:eptic}

\end{table*}

\begin{table*}[!t]
\centering
\small
  \scalebox{1}
  {
  \begin{tabular}{lccccc}
    \toprule
      & & & &BLEU  \\
      Model &  AL &  AP &  \scalebox{0.80}{Translation} & \scalebox{0.80}{Interpretation$^{ASR}$} & \scalebox{0.80}{Interpretation}  \\
    \midrule
    \addlinespace[0.3em]
    \textit{Europarl} \\
    train on \scalebox{0.8}{\textsf{<Source, Translation>}}&  0.61 &  2.84  &  22.78$^*$ & 11.47$^*$& 12.34$^*$   \\
    \hspace{1.5mm} adapt on \scalebox{0.8}{\textsf{<Source-FB, Raw>}}  &  \underline{0.61}&  \underline{2.76} & 20.71  &12.05 & 12.69 \\
    \hline
   \textit{Style transferred Europarl}\\
    train on \scalebox{0.8}{\textsf{<Source, Pseudo-I>}}\\
    \hspace{1.5mm} Seq2Seq (unsupervised) & 0.66 & 4.45 &  10.42  &7.79 & 10.33 \\
    \hspace{1.5mm} HPBMT (unsupervised) &  0.61&  2.92& 18.80 & 13.53 & 13.21  \\
    \hspace{1.5mm} PBMT (supervised) &  0.61 &  2.93 &  17.34  & 13.87 & 13.56   \\
    \hspace{1.5mm} HPBMT (supervised) & 0.62 & 3.00 &  18.55 & \textbf{14.26} & \textbf{13.60} \\
    \bottomrule
  \end{tabular}
  }
  \caption{Evaluation on human annotated Translation Test, Interpretation Test$^{ASR}$ and Interpretation Test. $^*$: performance gap. \underline{Underlined}: lowest delay across systems. \textbf{Bold}: Best BLEU on Interpretation Test. 
  }
  \label{tab:focus} 
\end{table*}

\subsection{Datasets}\label{sec:data}

We conducted evaluation investigation on four languages pairs, including German (DE), French (FR), Polish (PL), Italian (IT) $\rightarrow$ English (EN) , and used Europarl v7 corpus \cite{koehn2005europarl} for training a SiMT model for each pair (see Table~\ref{tab:eptic} for data statistics).
For DE-EN, our annotated test set has 1,051 triples, for \textsf{Interpretation$^{ASR}$} and \textsf{Interpretation}. For the rest, we used EPTIC \cite{bernardini2016epic}, a small-scale parallel corpus with data collected from the EP Plenary;  it has source languages of FR, PL and IT, with 675, 463 and 480 instances, respectively.

In the experiments of bridging the evaluation gap, \textsf{Raw} has 120,114 and 1,000 utterances for training and dev sets, while  \textsf{\textsf{Clean}} has 4,240 triples, all used for training style transfer models. To train PBMT, we augmented \textsf{Clean} by forward translating its source-side data to the target language, together with EPTIC, while using EPTIC to select the best weights for PBMT.
 We deployed the trained style transfer models on translations of Europarl (DE-EN) to get \textsf{Pseudo-I}. Pairing it with source sentences of Europarl gives us style transferred Europarl.
\subsection{Model}\label{sec:basemodels}
\paragraph{Baseline}
We used wait-k (with k=3) as SiMT systems for its simplicity and effectiveness \cite{DBLP:conf/acl/MaHXZLZZHLLWW19}. We compared the following wait-k baselines: i) trained on Europarl; ii) adapted on \textsf{Raw}. 
Performance was evaluated by BLEU\footnote{https://github.com/mjpost/sacreBLEU}, 
average proportion (AP) and lagging (AL) \cite{DBLP:journals/corr/ChoE16, DBLP:conf/acl/MaHXZLZZHLLWW19}.
AP measures the percentage of read source tokens for every generated target token, while AL measures the number of lagged source tokens until all source tokens are read.

\paragraph{Style Transfer Models}  In supervised settings, we used PBMT and HPBMT; in unsupervised settings we only used HPBMT, as PBMT requires additional paired data to find the best weights. 
We deployed Moses \citep{DBLP:conf/acl/KoehnHBCFBCSMZDBCH07} for above systems.~We also experimented with a Seq2Seq (unsupervised) model \cite{DBLP:conf/naacl/OttEBFGNGA19} to compare.
 

\subsection{Performance Gap}\label{sec:evalgap}

We train separate wait-k models for  the four language pairs and report the evaluation results on their corresponding Translation Test and Interpretation Test\footnote{A one-to-one correspondence exists between both sets for all language pairs.} in Table~\ref{tab:eptic}. The observed significant gap of up-to 13.83 BLEU score (24.47 vs 10.64 for IT)
highlights the daunting task SiMT models face in real-life SI. Interestingly, the gap for DE-EN is the lowest, and this is likely to be due to the fact that both are Germanic languages.


{We explored the feasibility of narrowing the performance gap using our T2I method on DE-EN. Being a head-final language, German is more difficult to interpret than head-initial languages (e.g., EN, FR, IT and PL}), and interpreters must hold information until verb phrases are heard~\cite{mizuno2017simultaneous}. Furthermore, having created the datasets for German, our experimental setup was year/domain-consistent for training the baselines and style-transfer models, which allows us to isolate if the improvement was purely achieved by our T2I transfer method.

Full results are reported in Table~\ref{tab:focus}.
%
When wait-k was adapted on \textsf{Source-FB, Raw}, the lowest delay was seen, implying using interpretation corpora is effective in reducing delay. Translation quality can be further boosted with our style transfer method, as discussed next.


\begin{table*}[!t]
\centering
\small
  \scalebox{0.9}
  {
  \begin{tabular}{p{3cm}|p{3cm}|p{3cm}|p{3cm}}
    \toprule
    \textbf{Source} & \textbf{Gold} & \textbf{wait-k} & \textbf{wait-k + T2I}  \\
    \midrule
    Es erfüllt mich mit großer Traurigkeit. &It is with great sadness. & I am with great sadness. & I'm very sorry about that.\\
    \hdashline
    Der Bericht begrüßt außerdem ausdrücklich den Vorschlag der Kommission für eine horizontale Richtlinie zum Thema Antidiskriminierung. &The report also explicitly welcomes the Commission’s proposal for a horizontal directive covering all forms of discrimination. & The report also expressly welcomes the Commission's proposal for a horizontal directive on the subject of anti-discrimination. & The report welcome the commission's proposal for a horizontal directive on anti-discrimination legislation. \\
    \bottomrule
  \end{tabular}
  }

  \caption{Examples of translation predicted by wait-k and translation predicted by a style transferred model, along with their source sentences and gold-translation.}

  \label{tab:demo}
\end{table*}

\subsection{Impacts of T2I Style Transfer}\label{sec:t2ianalysis}

\paragraph{Quantitative Analysis} 
Our approach yields significantly better results on \textsf{Interpretation$^{ASR}$} compared to baselines. Our best model outperformed pre-trained wait-k by 2.79 BLEU score. On \textsf{Interpretation}, we see a similar trend but with a smaller margin. We speculate the drop occurred because the T2I models were trained on ASR outputs, which is in the same domain as targets of \textsf{Interpretation$^{ASR}$}.

Nevertheless, all T2I models work consistently well in supervised and unsupervised settings. Moreover, our approach surpasses Seq2Seq by 6.47 points on \textsf{Interpretation$^{ASR}$}, verifying that in low-resource settings SMT is superior to NN. Our results, including adapting wait-k on \textsf{Raw} and using T2I to create training corpus, suggest that adequate numbers of paired translation, clean interpretation would lead to decreased delay and better translation quality.

The limitation, however, is that the BLEU score still remains relatively low,
which is not surprising, for we only used a minimal number of parallel data in the style transfer process. Hence, while our method does not remove the performance gap, it can still serve as a data augmentation technique to complement future interpretation training data. 


\paragraph{Qualitative Analysis}
To compare translations produced by the vanilla wait-k and its variants trained on T2I transferred data, we give examples in Table~\ref{tab:demo} along with their sources and gold translations. In the first example, T2I variant is colloquial, implying interpreters giving up the original words and restating the source message \citep{camayd2011cognitive}. T2I variant in the other example is a more condensed translation by dropping unimportant words \citep{DBLP:journals/corr/abs-2011-04845}, such as "also expressly" and "the subject of". Both examples confirm human interpreters' tactics~\citep{DBLP:conf/naacl/HeBD16}.

\section{Conclusion}

\label{sec:conclusion}
We investigated the SiMT evaluation gap when SiMT models were tested on interpretation vs translation, across four language pairs. To the best of our knowledge, this is the first work quantifying this gap empirically. To bridge the gap, we proposed a data augmentation style transfer technique 
to create parallel pseudo-interpretations from abundant offline translation data. Our results show an improvement of 2.8 BLEU score. We hope our work and the highlighted evaluation discrepancy can encourage further developments of datasets and models more reflective of real-world SI scenarios. 

\section{Acknowledgements}
This work is supported by the ARC Future Fellowship FT190100039 and an Amazon Research Award to G.H. We would like to thank anonymous reviewers for their valuable comments.

\bibliography{custom}
\bibliographystyle{acl_natbib}

\clearpage
\appendix
\section{Appendix}

\label{sec:appendix}

\subsection{Corpus Construction}
In this section, we will describe the process of collecting genuine data, and creating interpretation datasets with a proposed dynamic programming algorithm. We will then present a test set annotated by a bilingual annotator to ensure the genuineness of our experimental results and analysis.

\subsubsection{Data Collection}
 
We collected genuine data from the European Parliament (EP) Plenary where debates are carried out among representatives of member states of the European Union who speak their native languages; to facilitate communication, simultaneous translation services are provided. From 2008/09/01 to 2012/11/22,  source speeches transcriptions, post-edited offline translations and interpretation audios are available. We selected German-English as our target language pair. We used a number of heuristics methods to identify the nationality of the speakers and crawled the data accordingly.  
Note post-edited translations are only available during this period due to the changes of EP's policies, while audios and transcriptions are provided from 2004 till today. We will leave collecting and making use of the full data in our future work. In total, we downloaded 238 debates and 2415 video files in the mp4 format, with a total size of 500GB. We then used google speech API to transcribe and normalize ASR outputs, while using  offline translations to retrieve name entities. Thus, 19,368.24 minutes were transcribed, with a total cost of 697.257 USD.
\begin{table}[!ht]
    \small
   \scalebox{1}{
    \begin{tabular}{|p{7.1cm}|}
    \hline
        \textbf{Source}
        \begin{itemize}
            \item  Herr Prsident, liebe Kolleginnen und Kollegen! Wir haben im Ausschuss ber diesen Antrag lange debattiert, wir haben mit grofer Mehrheit eine Entscheidung getroffen, aber es hat gestern und heute eine Flle von Hinweisen und Anregungen gegeben, die sich vor allem auch deshalb ergeben haben, weil andere Ausschsse noch Beratungsgegenstnde hinzugefgt haben.
            \item Es scheint mir sinnvoll zu sein, nicht heute zu entscheiden, sondern noch einmal die Gelegenheit zu haben, eine Lsung zu finden, die dann auch das Parlament tragen kann. Deshalb bitte ich darum, die Verschiebung heute zu beschlieen. Danke.
        \end{itemize} \\
        \hline
        \textbf{Translation}
        \begin{itemize}
            \item Mr President, ladies and gentlemen, we debated this motion long and hard in the committee, and we reached a decision backed by a large majority, but yesterday and today, there has been an abundance of advice and suggestions that have come about primarily because other committees have added extra subjects for discussion.
            \item It seems to me that it would be a good idea not to make the decision today but, instead, to have the opportunity at a later date to find a solution which Parliament is then in a position to support. I therefore ask that you adopt this deferral today. Thank you.
        \end{itemize}\\
        \hline
        \textbf{Transcript}
        \begin{itemize}
            \item Mr. President dear colleagues in the committee, we discussed this motion at some length.
            \item We took a decision by a large majority, but between yesterday and today there have been a number of suggestions and indications that have Arisen because other committees. I've also been involved in this procedure.
            \item So we think it would it would be more intelligent not to take a decision on this report today but to give more time for us to try to find a solution to all of these issues that have been raised that all of Parliament can support. This is why I request that we decide on postponement today.
            \item Thank you .
        \end{itemize}\\
    \hline
    \end{tabular}}
    \caption{Example of the constructed dialogues.}
    \label{tab:example}
\end{table}

\subsubsection{Data Cleaning}


To build a high-quality dataset, we enacted a series of sophisticated automatic and manual pre-processing steps to filter and clean dialogues.  The total number of dialogues is 5,239. Initially, we removed dialogues whose source was non-German. To keep the essence of the debates, we filtered out below non-essential components and considered them as transitions between conversations: i) dialogues with less than 20 tokens; ii) dialogues which have pre-defined signals (e.g., "the vote will take place") and they have less than 100-150 words, depending on the signals. The number of data points removed is 2,174. Human investigation on disregarded dialogues confirmed these heuristics. We also discarded those whose source and translation have a different number of sentences, which was, however, rare. The above steps yielded a number of 1,872 data points. Following that, we manually deleted dialogues which were non-essential contents of the debates, while truncating transcriptions whose first and last sentences did not match the corresponding post-edited translation; most of deleted sentences were results of imperfect audio segmentation. After the manual process, 987 dialogues are retained, representing the essence of the debates. Table \ref{tab:example} is one example of the dialogues.

\subsubsection{Parallel Dataset Creation}
The procedure for constructing parallel interpretation data is described as follows. Firstly, we aligned sentences in the offline translation with those in the interpretation transcription. As the transcription may not be well segmented during the ASR process, we identified sentences in the transcript with stanza\footnote{\url{https:/ /github.com/stanfordnlp/stanza}}. Note each source sentence and post-edited translation sentence in the collected dialogues may comprise several sentences. We call them super-sentences and we don't perform sentence splitting on those super-sentences yet. Next, for each dialogue, we segmented the transcription sentences based on the number of super-sentences in the corresponding translation using dynamic programming, details of which will be discussed in the following section. This step is important due to the fact that unlike post-edited translations which tend to be long and formal, in real-life SI scenarios a source sentence (i.e., German in our case) can often broken into multiple smaller pieces. Hence, it is necessary to recognize and rejoin those pieces into chunks. We chose the candidate, i.e., segmented sentences in the interpretation transcription, which had the highest similarity score to the English translation in the semantic space\footnote{Similarity scores are calculated with \url{https://github.com/UKPLab/sentence-transformers/tree/master/sentence_transformers}}, as the output. More specifically, each of the chunks was semantically similar to one super-sentence in the translation. Since such a super-sentence in any dialogue corresponds to a long, formal German super-sentence, equally each of the chunks can be allocated to that source sentence. This gives us a super-sentence-level dataset, named \textsf{Super}, consisting of 3,683 triples <\cmss{source, translation, transcript}> . This step is done on English pairs, as we believe calculating similarity scores in the same language yields more accurate results than comparing the semantic similarity between different languages. 


Following that, we also tried to segment super-sentences with almost the identical procedure\footnote{The only difference here is that we used a multilingual encoder}. This gave us a sentence-level corpus, the filtered version of which is \textsf{Clean} in the main paper. After inspecting the outputs of the resulting dataset, we noticed it contained noises that were contributed by many factors, the most important of which of is occasional interpreting failure. Hence, we decided to recruit a bilingual German-English speaker to pair sentences manually and they become the test data in this work. 


\subsubsection{Sequence Segmentation/Alignment with Dynamic Programming}
We use a dynamic programming algorithm, as shown in Algorithm \ref{alg:npi}, to segment target utterances and perform alignment in the semantic space. As shown in Table \ref{tab:example}, each source sentence has its own correspondence of translation, so we only need to align segments of sentences to that translation sentence, in order to have a parallel source-interpretation corpus. Hence, we dynamically divide sentences in the transcription by the number of translation sentences, calculate the similarity score for each pair while considering the accumulated scores for sequences preceding it. We then traceback the candidate with the best score. The time complexity of this algorithm is $O(KN^2)$, where $K$ is the number of source sentences and $N$ is the number of target utterances. This algorithm is applicable to creating both the super-sentence-level and sentence-level parallel datasets. 

\newcommand{\vy}{\pmb{y}}
\newcommand{\vx}{\pmb{x}}
\newcommand{\vX}{\pmb{X}}
\newcommand{\vY}{\pmb{Y}}

\renewcommand{\algorithmicrequire}{\textbf{Input:}}
\renewcommand{\algorithmicensure}{\textbf{Output:}}

\begin{algorithm}[!t]
\caption{Constrained segmentation}
\begin{algorithmic}[1]
\REQUIRE{$\vY$: List of unsegmented target utterances, $N:$ Length of target sequence, $\vX:$ List of segmented source, $K:$ Number of source segments, $d$: A distance similarity metric.}
\ENSURE{$T$: The DP table with optimal scores}
\STATE // initialisation
\FOR{$i = 1 ... {N}$}
\STATE $T_{1,i} = d(\vX_1,\vY_{1:i})$
\ENDFOR
\STATE // Filling out $T$ based on the DP relation
\FOR{$k = 2 ... K$}
\FOR{$i = k ... N$}
\STATE $T_{k,i} = \max\limits_{k \le j \le i} \left[T_{k-1,j-1} + d(\vX_k, \vY_{j:i})\right]$ 
\ENDFOR 
\ENDFOR
\RETURN $T$
\end{algorithmic}
\label{alg:npi}
\end{algorithm}

\subsection{Europarl vs Style Transferred Europarl} 
To illustrate the outcomes of style transfer models, we provides illustrations of target sentences in Europarl and style transferred Europarl (\textsf{Pseudo-I}) in Table \ref{tab:ST_demo}. The first example involves word reordering and word replacement that "so" is replaced with "therefore" before being put at front; it also involving word replacement in that "prepare its programme" is changed to "draw up the program" and "become familiar with it" changed to "be aware that". In the second example, "endorse" is replace with a common phrase "agree with".

\begin{table}[!t]
\centering
\small
  \scalebox{0.94}
  {
  \begin{tabular}{p{3.65cm}|p{3.67cm}}
    \toprule
    \textbf{Europarl} & \textbf{Style~Transferred~Europarl}  \\
    \midrule
    There has therefore been enough time for the Commission to prepare its programme and for us to become familiar with it and explain it to our citizens.& So there has been enough time for the Commission to draw up the program and for us to be aware that and explain it to our citizens. \\
    \hdashline
    I would urge you to endorse this. & I would ask you to agree with that. \\
    \bottomrule
  \end{tabular}
  }
  \vspace{-2mm}
  \caption{Examples of Europarl vs Style Transferred Europarl.}
  \vspace{-2.8mm}
  \label{tab:ST_demo}
\end{table}

\subsection{Discussion}
\subsubsection{Analysis on N-grams.}

\pgfplotstableread[row sep=\\,col sep=&]{
    method & adapt & seq2seq & HPBMT(unsup) & PBMT & HPBMT\\
1-gram & 20.71 & 15.95 & 21.83 & 19.59 & 19.52\\
2-gram & 11.14 & 10.48 & 11.34 & 12.13 & 12.16 \\
3-gram & 06.05 & 06.16 & 06.01 & 07.19 & 07.64 \\
4-gram & 03.16 & 03.42 & 03.24 & 04.25 & 04.40 \\
    }\interngram

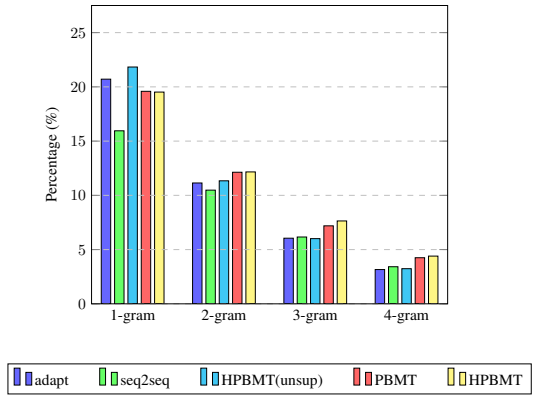
\begin{figure}[!t]
\centering 
  \begin{tikzpicture}[thick,scale=0.73, every node/.style={scale=0.73}]
      \begin{axis}[
            ybar, axis on top,
            title style={at={(0.5,0.97)},anchor=north,yshift=-0.1, draw=gray},
            height=7cm, 
            width=8cm,
            bar width=0.17cm,
            ymajorgrids, tick align=inside,
            major grid style=dashed,
            enlarge y limits={value=.1,upper},
            ymin=0, ymax=25,
            x tick style={draw=none},
            enlarge x limits=0.15,
            legend style={
                at={(0.5,-0.2)},
                anchor=north,
                legend columns=-1,
                /tikz/every even column/.append style={column sep=0.5cm}
            },
            ylabel={Percentage (\%)},
            ylabel near ticks,
            symbolic x coords={
               1-gram, 2-gram, 3-gram, 4-gram},
           xtick=data,
        ]
        \legend{adapt, seq2seq, HPBMT(unsup), PBMT, HPBMT}
        \addplot[fill=blue!60] table[x=method,y=adapt]{\interngram};
        \addplot[fill=green!60] table[x=method,y=seq2seq]{\interngram};
        \addplot[fill=cyan!60] table[x=method,y=HPBMT(unsup)]{\interngram};
        \addplot[fill=red!60] table[x=method,y=PBMT]{\interngram};
        \addplot[fill=yellow!60] table[x=method,y=HPBMT]{\interngram};
        
      \end{axis}
      \end{tikzpicture}

\caption{Percentage of introduced correct n-grams.}
\label{fig:ngram}
\end{figure}
To investigate what led to the improvement, we first computed n-grams present in gold \textsf{Interpretation} but not in outputs predicted by the baseline wait-k. Then we examined the amount of n-grams newly introduced by each model that are overlapped with the n-grams calculated previously. As shown\footnote{We dropped unsupervised, supervised for the sake of clarity of the plot, while just using HPBMT(unsup) to indicate HPBMT(unsupervised).} in Figure \ref{fig:ngram}, PBMT and HPBMT in supervised settings consistently introduce more n-grams than others with the only exception that adapt and HPBMT(unsup) produce more new 1-gram. Essentially, this implies that style transferred Europarl has effectively captured more interpretation features than the original Europarl.



\end{document}